\pdfoutput=1

\documentclass[11pt]{article}

\usepackage[preprint]{acl}

\usepackage{times}
\usepackage{latexsym}

\usepackage[T1]{fontenc}

\usepackage[utf8]{inputenc}

\usepackage{microtype}

\usepackage{inconsolata}

\usepackage{graphicx}

\usepackage{amsfonts}
\usepackage{mathtools}

\usepackage{algorithm}
\usepackage{algorithmicx}
\usepackage{algcompatible}
\usepackage{xcolor}
\usepackage{multirow}
\usepackage{colortbl}
\usepackage{adjustbox}
\usepackage{tabularx, booktabs}
\usepackage{caption}
\usepackage{subcaption}
\usepackage{tcolorbox}
\usepackage{enumitem}

\usepackage{pifont}

\usepackage{amsmath}
\usepackage{xspace}
\usepackage{makecell}
\usepackage{subcaption}

\usepackage{amsthm}

\usepackage{graphicx}
\usepackage{subcaption}
\usepackage{float}

\usepackage{booktabs}
\usepackage{siunitx}
\usepackage{listings}

\definecolor{col}{HTML}{598BE7}
\definecolor{col2}{HTML}{F54254}
\definecolor{col3}{RGB}{66, 244, 133}
\definecolor{RED}{RGB}{241, 79, 33}
\definecolor{GREEN}{RGB}{126, 185, 0}
\definecolor{BLUE}{RGB}{0, 163, 238}
\definecolor{YELLOW}{RGB}{254, 184, 0}
\definecolor{GRAY}{RGB}{114, 114, 114}
\definecolor{POSTECH}{RGB}{200, 1, 80}

\usetikzlibrary{patterns}
\newcolumntype{R}[1]{>{\raggedleft\arraybackslash}p{#1}}
\newcolumntype{L}[1]{>{\raggedright\arraybackslash}p{#1}}

\hypersetup{
  colorlinks=true,
  linkcolor=col,   
  citecolor=col,    
  urlcolor=col    
}

\lstdefinestyle{pythonstyle}{
    language=Python,           
    basicstyle=\ttfamily\small,
    keywordstyle=\color{black}, 
    stringstyle=\color{orange!60!black},   
    commentstyle=\color{green!50!black}, 
    numbers=left,              
    numberstyle=\tiny\color{gray}, 
    breaklines=true,           
    frame=single,              
    rulecolor=\color{black},   
    showstringspaces=false,    
    tabsize=4,                 
    captionpos=b,              
    breakatwhitespace=false,   
}

\lstdefinestyle{pythonstyle2}{
  language=Python,
  basicstyle=\ttfamily\tiny,
  keywordstyle=\color{blue}\bfseries,
  stringstyle=\color{orange!60!black},
  commentstyle=\color{green!50!black},
  numbers=none,
  breaklines=true,
  frame=none,
  showstringspaces=false,
  columns=fullflexible,
  xleftmargin = -12pt,
  xrightmargin = -12pt,
  framesep=0pt,
  framexleftmargin=-10pt,
  framexrightmargin=-10pt,
  aboveskip = -5pt,
  belowskip = -5pt,
  lineskip = -3pt,
  escapeinside={(@}{@)},
}

\lstdefinestyle{pythonstyle3}{
  backgroundcolor=\color{col!5},
  basicstyle=\ttfamily\footnotesize,
  keywordstyle=\color{black},
  stringstyle=\color{orange!60!black},
  commentstyle=\color{green!50!black},
  numbers=none,
  breaklines=true,
  frame=none,
  showstringspaces=false,
  columns=fullflexible,
  keepspaces=true,                 
  showspaces=false,
  framesep=0pt,
  aboveskip = 0pt,
  belowskip = 0pt,
  escapeinside={(@}{@)},
}

\lstdefinestyle{pythonstyle4}{
  backgroundcolor=\color{col!5},
  language=Python,
  basicstyle=\ttfamily\footnotesize,
  keywordstyle=\color{blue}\bfseries,
  stringstyle=\color{orange!60!black},
  commentstyle=\color{green!50!black},
  identifierstyle=\color{black},
  numbers=none,
  breaklines=true,
  frame=none,
  showstringspaces=false,
  columns=fullflexible,
  keepspaces=true,                 
  showspaces=false,
  framesep=0pt,
  aboveskip = 0pt,
  belowskip = 0pt,
  escapeinside={(@}{@)},
}

\title{Confidence-Gated Transductive Test Generation for Code Reranking}
\author{
Sungjae Lee$^1$, Youngsik Yoon$^1$, Seockbean Song$^2$, Siwei Wang$^3$, Wei Chen$^3$, Jungseul Ok$^{1,2}$\thanks{Corresponding author.}\\
$^1$Department of Computer Science and Engineering, POSTECH, South Korea\\
$^2$Graduate School of Artificial Intelligence, POSTECH, South Korea\\
$^3$Microsoft Research Asia, China\\
\{sungjaelee25, ysyoon97, shinebobo, jungseul.ok\}@postech.ac.kr, \\
\{siweiwang, weic\}@microsoft.com
}

\begin{document}
\maketitle

\begin{abstract}

Test case synthesis is crucial for evaluating and ranking programs generated by large language models (LLMs). However, constructing high-quality test cases remains challenging because reliable expected outputs are often difficult to obtain. We propose Confidence-Gated Transductive Test Generation (CoTT), which first uses an efficient inductive procedure and invokes transductive generation only when inductive confidence is low. This adaptive design improves output reliability while allocating extra computation only when needed. On code reranking benchmarks, CoTT outperforms prior baselines across the reported metrics while reducing cost relative to applying transductive generation to every input. These results show that confidence-based allocation of test-time computation provides a favorable efficiency--effectiveness trade-off with a single efficient LLM.
\end{abstract}

\section{Introduction}

\begin{figure*}[!t]
    \centering
    \includegraphics[width=0.97\textwidth, trim={0 0 0 0},clip]{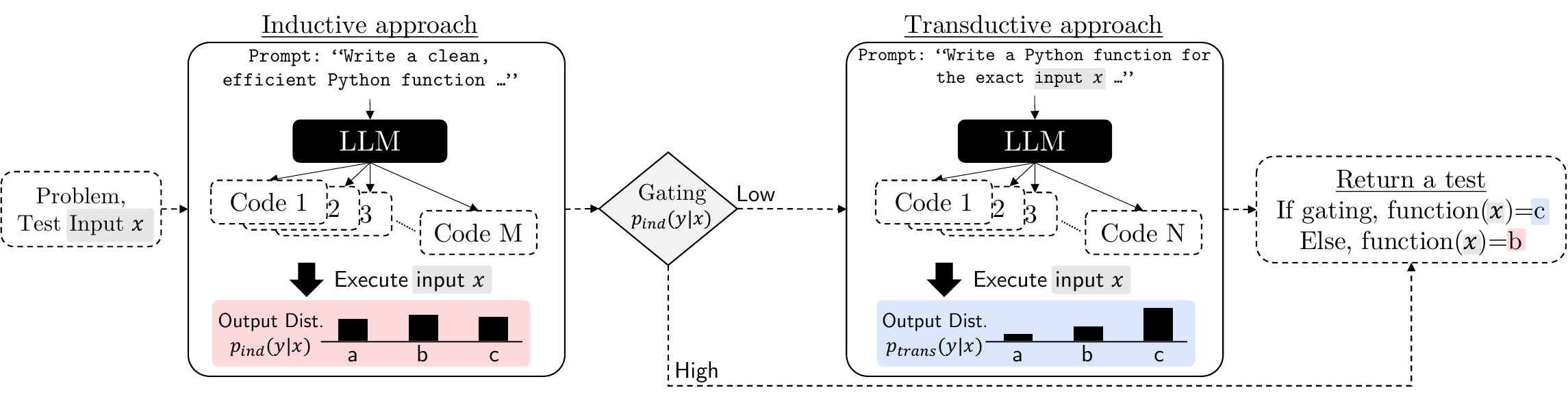}
    \vspace{-0.3cm}
    \caption{\textbf{Overview of CoTT.}
    CoTT predicts the expected output for a synthesized test input using inductive generation, and invokes transductive generation only when inductive confidence is low. This confidence-gated design improves output reliability while allocating additional computation only when needed.
    }\label{fig:concept}
    \vspace{-0.5cm}
\end{figure*}

Test cases are a fundamental tool for evaluating and scoring code \citep{chen2021evaluating, austin2021program, chen2022codet,  yu2024reasoning}. With the rapid progress of large language models (LLMs) in code generation, test case synthesis has become an increasingly important component for providing execution-based signals to identify better program candidates \citep{ficek2025scoring}. In practice, this need is especially pronounced when high-quality human-written tests are sparse or unavailable, making synthesized tests a practical source of execution-based feedback \citep{ni2023lever}. However, most prior work has focused primarily on improving code generation, often relying either on a limited number of ground-truth tests or on a small set of simply generated tests \citep{zhong2024debug, yu2024reasoning}.

One line of work executes multiple programs generated by an efficient LLM and uses agreement or consensus among them to validate synthesized tests \citep{chen2022codet, chen2024divide}. However, when the generated programs themselves are unreliable, such signals may still be insufficient for constructing synthesized test cases with reliable expected outputs. 
Another line of work uses input-conditioned generation or verifier-guided execution to reason about specific target inputs and assess candidate outputs  \citep{lin2025learning, lee2025program}. While effective, these approaches may require additional training, search, or auxiliary-model computation.

Constructing synthesized test cases with reliable expected outputs using a single efficient LLM, instead of relying on larger and more expensive LLMs, raises two key challenges. 
First, improving test quality requires more than simply scaling parallel sampling in the same way \citep{li2025s}; when expected-output estimates are unreliable, a more accurate mechanism is needed to refine them through additional computation. 
Second, this stronger mechanism should be triggered only when necessary, so that additional computation is allocated efficiently \citep{li2024escape, lee2025semantic}. In this sense, test construction requires both better expected outputs and deciding when extra computation is worthwhile.

In this work, we propose Confidence-Gated Transductive Test Generation (CoTT), an adaptive method for constructing synthesized test cases with reliable expected outputs using a single efficient LLM as described in Figure~\ref{fig:concept}. Rather than uniformly scaling programs, CoTT allocates additional computation only when stronger expected-output estimation is needed. CoTT combines an \emph{inductive} procedure for expected-output prediction with a \emph{transductive} procedure for input-specific program generation, invoking the latter only when inductive output agreement indicates low confidence. This design remains practical in a single-model setting and improves the reliability of synthesized tests for code reranking. Experiments across datasets show that CoTT improves the expected-output reliability of synthesized tests, leading to better code reranking with a single efficient LLM.

\section{Related Work}

\paragraph{Synthetic verification and scoring for code.}
Recent work uses synthetic tests, execution feedback, and learned critics to score or rerank code candidates produced by LLMs \citep{zeng2025acecoder, ma2025dynamic, yu2024reasoning, chen2022codet, chen2024divide}. These methods highlight synthesized verification as a key component of downstream code selection, especially when high-quality human-written tests are limited. Recent studies further argue that synthetic verification should be treated as an important evaluation problem for code reranking \citep{ficek2025scoring}. 
Our work is most closely related to this line, but treats synthesized test reliability as a key bottleneck for code reranking and evaluates it directly on reranking benchmarks.

\paragraph{Test synthesis and the oracle problem.}
A core challenge in test synthesis is the \emph{oracle problem}: reliable expected outputs are often harder to obtain than test inputs themselves. Existing approaches to this problem largely fall into two categories. One line of work estimates expected outputs by executing many general solution candidates and aggregating their outputs through agreement or majority voting \citep{chen2022codet, chen2024divide, liu2025rstar}. 
Another line of work relies on stronger or input-conditioned procedures to reason about a target input, or to select and verify candidate outputs with stronger external models \citep{lee2025program, lin2025learning}. 
While effective, these approaches are tailored either to settings where execution-based consensus is sufficiently reliable or to settings with access to stronger auxiliary models.

\section{Method}\label{sec:method}

\begin{table*}[t!]
\centering
\begin{adjustbox}{width=1.0\textwidth}
\begin{tabular}{lccccccc}
\toprule 
\multirow{2.5}{*}{Test Output Generation Method} 
& \multicolumn{3}{c}{HumanEval-R+} 
& \multicolumn{3}{c}{MBPP-R+} 
& \multirow{2.5}{*}{Avg. Cost $\downarrow$} \\
\cmidrule{2-4} \cmidrule{5-7}
& Top-1 $\uparrow$ & Bottom-1 $\uparrow$ & Spearman $\uparrow$ 
& Top-1 $\uparrow$ & Bottom-1 $\uparrow$ & Spearman $\uparrow$ 
& \\ 
\midrule
CoT \cite{wei2022chain} 
& 42.5 & 48.3 & 0.440 
& 42.0 & 43.9 & 0.349 
& 0.060 \\
SOL-VER \cite{lin2025learning} 
& 41.1 & 52.1 & 0.464 
& 42.8 & 47.1 & 0.395 
& 0.080 \\
SYNTRA \cite{lee2025program}  
& 56.6 & 59.8 & 0.585 
& 51.8 & 56.7 & 0.539 
& 0.069 \\
rStar-Coder \cite{liu2025rstar}  
& 59.8 & 65.4 & 0.611 
& \underline{59.8} & \underline{58.1} & 0.559 
& 0.058 \\
\cellcolor[HTML]{EFEFEF}TT w/o Gating (Ours) 
& \cellcolor[HTML]{EFEFEF}\textbf{63.1} 
& \cellcolor[HTML]{EFEFEF}\underline{66.2} 
& \cellcolor[HTML]{EFEFEF}\textbf{0.676} 
& \cellcolor[HTML]{EFEFEF}58.5 
& \cellcolor[HTML]{EFEFEF}56.1 
& \cellcolor[HTML]{EFEFEF}\underline{0.576} 
& \cellcolor[HTML]{EFEFEF}0.077 \\
\cellcolor[HTML]{EFEFEF}CoTT (Ours) 
& \cellcolor[HTML]{EFEFEF}\underline{62.5} 
& \cellcolor[HTML]{EFEFEF}\textbf{66.7} 
& \cellcolor[HTML]{EFEFEF}\underline{0.650} 
& \cellcolor[HTML]{EFEFEF}\textbf{61.6} 
& \cellcolor[HTML]{EFEFEF}\textbf{58.7} 
& \cellcolor[HTML]{EFEFEF}\textbf{0.580} 
& \cellcolor[HTML]{EFEFEF}0.065 \\
\bottomrule
\end{tabular}
\end{adjustbox}
\caption{
Comparison of reranking performance and average cost for test output generation methods on HumanEval-R+ and MBPP-R+. Metrics include Top-1 accuracy, Bottom-1 accuracy, Spearman’s $\rho$, and average cost (\$). \textbf{Bold} values denote the best results, and \underline{underlined} values indicate the second-best.
}
\label{tab:main-rank}
\vspace{-4mm}
\end{table*}

We propose \emph{Confidence-Gated Transductive Test Generation} (CoTT), an adaptive framework for constructing high-quality test cases with reliable expected outputs using a single efficient LLM. Given a synthesized test input $x$, CoTT first predicts its expected output with an efficient \emph{inductive} procedure, then invokes a more expensive \emph{transductive} procedure only when the inductive estimate is uncertain. As illustrated in Figure~\ref{fig:concept}, CoTT uses the inductive prediction for high-confidence inputs and switches to a transductive prediction only for low-confidence ones. The resulting expected outputs are used to construct test cases for downstream code scoring and reranking.

\paragraph{Inductive output distribution.}
For each synthesized test input $x$, we first estimate an inductive output distribution by sampling a pool of general solution candidates for the original problem with instruction \texttt{``Write a clean, efficient Python function ...''} as shown in Figure~\ref{fig:appendix-prompt-induc} and executing them on $x$. Let $n_{\mathrm{ind}}(y;x)$ denote the number of inductive samples whose execution on $x$ returns output $y$. We define the empirical inductive distribution as
\begin{equation}
p_{\mathrm{ind}}(y \mid x)
=
\frac{n_{\mathrm{ind}}(y;x)}{\sum_{y'} n_{\mathrm{ind}}(y';x)}.
\end{equation}
The corresponding inductive confidence is
\begin{equation}
c(x)=\max_y p_{\mathrm{ind}}(y \mid x).
\end{equation}
Intuitively, $c(x)$ is the majority ratio of the most frequent inductive output. When this value is high, the set of general solution candidates already provides a reliable estimate of the expected output.

\paragraph{Confidence-gated transductive output distribution.}
When the inductive confidence is low, CoTT triggers a transductive procedure specialized to the target input $x$ using the instruction \texttt{``Write a Python function for the exact input ...''} as shown in Figure~\ref{fig:appendix-prompt-transduc}. In this step, the model generates instance-specific programs conditioned on both the problem and the particular target call, executes them on $x$, and forms a transductive output distribution
\begin{equation}
p_{\mathrm{trans}}(y \mid x)
=
\frac{n_{\mathrm{trans}}(y;x)}{\sum_{y'} n_{\mathrm{trans}}(y';x)},
\end{equation}
where $n_{\mathrm{trans}}(y;x)$ counts how many transductive samples produce output $y$ on $x$.

Given a confidence threshold $\tau$, CoTT selects the final expected output by
\begin{equation}
\hat{y}(x)=
\begin{cases}
\arg\max_y p_{\mathrm{ind}}(y \mid x), & \text{if } c(x)\ge \tau,\\[4pt]
\arg\max_y p_{\mathrm{trans}}(y \mid x), & \text{if } c(x)<\tau.
\end{cases}
\end{equation}
Thus, CoTT does not combine inductive and transductive distributions; instead, confidence gating selects which distribution to trust for each test input, as illustrated in Figure~\ref{fig:concept} and Algorithm~\ref{alg:cott}. Given a threshold $\tau$, the inductive prediction is used by default, and transductive generation is invoked only when the inductive confidence is low. In all main experiments, we use a threshold $\tau=0.8$ across datasets; hyperparameter settings are provided in Table~\ref{appendix:tab:hyperparameter}. Each synthesized test input $x$ is paired with its predicted output $\hat{y}(x)$ to form a test case $(x,\hat{y}(x))$ for executing and scoring candidate programs in reranking based on pass rate.

\section{Experiments}

We evaluate CoTT on full datasets of HumanEval-R+ and MBPP-R+ with Llama-3.1-8B-Instruct. Following \citet{ficek2025scoring}, these are code reranking benchmarks derived from HumanEval+ and MBPP+ \citep{liu2023your}, where each task is paired with five candidate solutions of varying correctness and the goal is to recover the ground-truth ranking induced by the reference test suites. Our focus is on \emph{test output generation}: given a shared set of synthesized inputs, each method predicts the corresponding outputs to form test cases, executes all candidate solutions on them, and ranks the candidates by their pass rate on these test cases.

We report Top-1 accuracy, Bottom-1 accuracy, and Spearman’s $\rho$ following \citet{ficek2025scoring}, where the first two measure whether the highest- and lowest-ranked candidates match the reference ranking, and Spearman’s $\rho$ measures correlation with the full reference ordering. We report the average per-problem cost over both benchmarks, computed from all input and output LLM tokens used for synthesized test construction, including both test-input generation and expected-output generation.\footnote{Prices for Llama-3.1-8B-Instruct (\$0.10 per 1M input tokens and \$0.10 per 1M output tokens) were obtained from \url{https://artificialanalysis.ai/models/llama-3-1-instruct-8b} (accessed March 16, 2026).} Unless otherwise specified, all methods use the same synthesized test inputs, so comparisons isolate differences in test output generation. We construct these inputs to be challenging and discriminative by selecting high-disagreement seeds from temporary program executions and expanding them with type-aware mutation, without using ground-truth solutions.

\paragraph{Baselines.}
We compare CoTT with representative transductive and inductive baselines for test output generation. \textbf{CoT} \citep{wei2022chain} predicts the expected output for each input with a single chain-of-thought pass. \textbf{SOL-VER} \citep{lin2025learning} and \textbf{SYNTRA} \citep{lee2025program} are transductive baselines that perform input-conditioned output generation with multiple CoT samples or output selection among candidates. \textbf{rStar-Coder} \citep{liu2025rstar} is our inductive baseline: it samples multiple general-purpose programs, executes them on each synthesized input, and predicts the expected output by majority vote. In our terminology, this corresponds to the \emph{inductive} approach, since it derives outputs from general candidate programs rather than input-specific reasoning. For baselines that include training or downstream optimization components, we apply only their test-output generation component under the same reranking pipeline. \textbf{TT w/o Gating} is an always-transductive variant that invokes
transductive generation for every synthesized test input and predicts its
expected output using $\arg\max_y p_{\mathrm{trans}}(y \mid x)$.
This variant isolates the effect of confidence gating.

\paragraph{Experimental Results.}
Table~\ref{tab:main-rank} shows that our methods achieve strong reranking performance relative to prior baselines.
On HumanEval-R+, TT w/o Gating obtains the highest Top-1 accuracy and Spearman's $\rho$, whereas CoTT obtains the highest Bottom-1 accuracy.
On MBPP-R+, CoTT achieves the best results across three reranking metrics.
Compared with TT w/o Gating, CoTT reduces the average inference cost from \$0.077 to \$0.065, while preserving most of its benefit on HumanEval-R+ and performing better on MBPP-R+.
Thus, confidence gating does not uniformly outperform always-transductive generation, but provides a favorable efficiency--effectiveness trade-off.
We additionally conduct a sensitivity analysis by sweeping the confidence threshold $\tau$ on both benchmarks; the results are provided in Appendix~\ref{sec:app_threshold_sensitivity}.
Additional analyses of alternative confidence scores and per-problem cost variation are provided in Appendices~\ref{sec:app_confidence_score} and \ref{sec:app_cost_statistics}.

\section{Analysis: When Does CoTT Help?}\label{sec:exp:analysis}

\begin{figure}[t]
    \centering
    \includegraphics[width=\columnwidth]
    {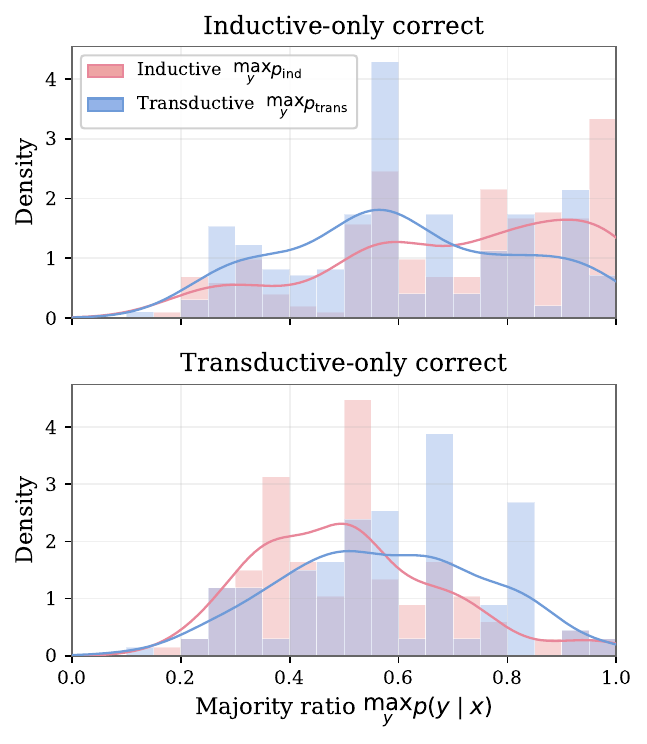}
    \vspace{-10mm}
    \caption{
    Comparison of the inductive and transductive majority-ratio distributions over synthesized test inputs for which the two predictions disagree. The top panel contains inputs for which only the inductive prediction is correct, whereas the bottom panel contains inputs for which only the transductive prediction is correct. The distributions correspond to $\max_y p_{\mathrm{ind}}(y\mid x)$ and $\max_y p_{\mathrm{trans}}(y\mid x)$, respectively.
    }
    \label{fig:gating_hist}
    \vspace{-7mm}
\end{figure}

\paragraph{Effectiveness of Confidence Gating.}
We first analyze why confidence gating improves the efficiency--accuracy trade-off. Recall that CoTT uses the inductive confidence
$
c(x) = \max_y p_{\mathrm{ind}}(y \!\mid\!\! x)
$
to decide whether to trust $\arg\max_y p_{\mathrm{ind}}(y\mid x)$ or invoke transductive refinement and use $\arg\max_y p_{\mathrm{trans}}(y\mid x)$. Figure~\ref{fig:gating_hist} shows that when only the inductive prediction is correct, $c(x)$ is typically high, so $p_{\mathrm{ind}}$ is already reliable and transduction is often unnecessary. In contrast, when only the transductive prediction is correct, $c(x)$ is much lower, indicating that $p_{\mathrm{ind}}$ is unreliable and that switching to $p_{\mathrm{trans}}$ is beneficial. This supports the inductive majority ratio as an effective trigger for allocating extra computation only when needed.

\paragraph{Case Study: When Transduction Helps.}
Consider the MBPP problem \texttt{``Write a function to find the Eulerian number a(n, m).''} For the synthesized input \texttt{eulerian\_num(0, 10)}, whose correct output is \texttt{0}, the inductive prediction is \texttt{1} while the transductive prediction is \texttt{0}. Although neither generated program is correct in general, the transductive program adds an input-specific boundary-case condition that correctly handles \texttt{n=0}, whereas the inductive program does not. Thus, even an imperfect transductive program can provide a useful complementary signal by exposing a target-specific error that the inductive prediction alone would miss. This illustrates how $p_{\mathrm{trans}}$ can recover the correct expected output even when general inductive solutions are weak. The corresponding prompts and generations are shown in Figures~\ref{fig:appendix-prompt-induc}, \ref{fig:appendix-prompt-transduc}, \ref{fig:appendix-induc-gen-1}, and \ref{fig:appendix-transduc-gen-1}.

\paragraph{Case Study: When Induction Suffices.}
We also observe the opposite pattern. For the MBPP problem \texttt{``Write a python function to find quotient of two numbers (rounded down to the nearest integer).''} and synthesized input \texttt{find(10, -2)}, the inductive prediction is correct, whereas the transductive prediction is not. Here the inductive samples consistently recover the correct rule \texttt{dividend // divisor}, while the transductive program unnecessarily introduces a negative sign, producing a rule like \texttt{a // -b}. The inductive executions are consistent on this input, yielding $c(x)=1.0$. This means that CoTT would retain $\arg\max_y p_{\mathrm{ind}}(y\mid x)$ rather than invoking transduction. This example shows that confidence gating preserves the efficient inductive path when the inductive prediction is reliable. The corresponding generations are shown in Figures~\ref{fig:appendix-induc-gen-2} and \ref{fig:appendix-transduc-gen-2}.

\section{Conclusion}

We propose CoTT, an adaptive framework for constructing synthesized test cases for code evaluation and reranking with a single efficient LLM.
CoTT selectively invokes transductive generation only when inductive confidence is low, allocating additional computation to uncertain test inputs.
Across both datasets, CoTT provides a favorable efficiency--effectiveness trade-off by maintaining strong reranking performance at a lower cost than always-transductive generation.
An interesting direction for future work is to co-evolve code and test generation using CoTT’s constructive feedback to iteratively improve both programs and tests.

\section*{Limitations}

Our study has several limitations. First, although CoTT does not rely on larger external LLMs, it may further benefit from external information sources, such as relevant code documentation or retrieval-augmented context, when available. Second, we do not explore integrating CoTT with code generation to develop a co-evolving code-and-test generation mechanism. Higher-quality synthesized tests may provide useful feedback for iterative code improvement in such a setting. Third, we evaluate CoTT using a single efficient LLM on two function-level Python code-reranking benchmarks. This controlled setting isolates test-oracle generation, but the results do not establish cross-model robustness or generalization to more realistic repository- or project-level tasks. Fourth, CoTT targets sandboxed code-reranking settings in which generated programs are executed under timeouts. The method may therefore be less applicable when code execution is limited, unavailable, or computationally expensive.

\section*{Acknowledgements}

This work was supported by the Institute of Information \& Communications Technology Planning \& Evaluation (IITP) grants funded by the Korea government (MSIT) (RS-2019-II191906, Artificial Intelligence Graduate School Program (POSTECH); RS-2024-00436680, Global Research Support Program in the Digital Field; RS-2024-00509258, Global AI Frontier Lab). This work was also supported by the IITP (Institute of Information \& Communications Technology Planning \& Evaluation)-ITRC (Information Technology Research Center) grant funded by the Korea government (Ministry of Science and ICT) (IITP-2026-RS-2024-00437866). This project was also supported by Microsoft Research Asia.
\vspace*{.8cm}

\section*{AI Assistants}

We used large language models solely to improve the manuscript’s clarity, grammar, and style, within the scope of "Assistance purely with the language of the paper". All scientific claims, experimental designs, and empirical results reported in this paper are the original work of the authors.



\bibliography{custom}

\clearpage

\appendix

\setcounter{algorithm}{0}
\renewcommand{\thealgorithm}{A\arabic{algorithm}}

\setcounter{table}{0}
\renewcommand{\thetable}{A\arabic{table}}

\newpage
\section*{Appendix}\label{sec:appendix}
\renewcommand{\thefigure}{A\arabic{figure}}
\setcounter{figure}{0}

\section{CoTT Algorithm}
Algorithm~\ref{alg:cott} presents CoTT and its key components: the inductive approach, confidence gating, and the transductive approach for constructing reliable test outputs.

\begin{algorithm}[htbp]
\caption{Confidence-Gated Transductive Test Generation (CoTT)}
\begin{algorithmic}[1]
\REQUIRE Problem prompt $p$, synthesized test inputs $\mathcal{T}$, LLM $\mathcal{M}$, number of samples $N_{\mathrm{ind}}, N_{\mathrm{trans}}$, threshold $\tau$
\ENSURE Synthesized test cases $\hat{\mathcal{T}}=\{(x,\hat{y}(x)) : x \in \mathcal{T}\}$

\STATE Sample inductive programs $\{g_j\}_{j=1}^{N_{\mathrm{ind}}} \leftarrow \mathcal{M}(p)$
\FOR{each test input $x \in \mathcal{T}$}
    \STATE Execute $\{g_j\}$ on $x$ and form $p_{\mathrm{ind}}(y \mid x)$ from output frequencies
    \STATE Compute inductive confidence $c(x)=\max_y p_{\mathrm{ind}}(y \mid x)$
    \IF{$c(x) \ge \tau$}
        \STATE $\hat{y}(x) \leftarrow \arg\max_y p_{\mathrm{ind}}(y \mid x)$
    \ELSE
        \STATE Sample transductive programs $\{h_k\}_{k=1}^{N_{\mathrm{trans}}} \leftarrow \mathcal{M}(p,x)$
        \STATE Execute $\{h_k\}$ on $x$ and form $p_{\mathrm{trans}}(y \mid x)$ from output frequencies
        \STATE $\hat{y}(x) \leftarrow \arg\max_y p_{\mathrm{trans}}(y \mid x)$
    \ENDIF
\ENDFOR
\STATE \textbf{return} $\hat{\mathcal{T}}=\{(x,\hat{y}(x)) : x \in \mathcal{T}\}$
\end{algorithmic}
\label{alg:cott}
\end{algorithm}

\section{Experimental Settings}
The hyperparameters used in our experiments are listed in Table~\ref{appendix:tab:hyperparameter}. 

\begin{table}[htbp]
    \centering
    \small
    \setlength{\tabcolsep}{8pt}
    \renewcommand{\thetable}{A\arabic{table}}
    \setcounter{table}{0}
    \resizebox{\linewidth}{!}{%
    \begin{tabular}{lc}
        \toprule
        Hyperparameter & Value \\
        \midrule
        \multicolumn{2}{c}{Test Construction and Code Generation} \\
        Number of test cases per problem & 10 \\
        Number of inductive code samples & 20 \\
        Number of transductive code samples & 10 \\
        Timeout per test case (seconds) & 3 \\
        Threshold for confidence gating & 0.8 \\
        \midrule
        \multicolumn{2}{c}{LLM Settings} \\
        Maximum context length (tokens) & 18{,}000 \\
        Maximum generated tokens per round & 1{,}500 \\
        Temperature & 0.5 \\
        \bottomrule
    \end{tabular}
    }
    \caption{Default hyperparameter settings used in our experiments. The hyperparameters are grouped into test construction and code generation settings, and LLM settings.}
    \label{appendix:tab:hyperparameter}
\end{table}

\section{Prompt}
\label{sec:app_prompt}
For code generation in the inductive approach, we follow the prompting style of \citet{yu2024reasoning}. Specifically, the model is prompted to generate a general-purpose solution to the original programming problem.
For code generation in the transductive approach, we use a modified version of this prompt that is adapted to our setting.
Instead of generating a solution intended to generalize broadly, the model is instructed to produce code specialized to the given target input.
The full prompts used for inductive and transductive code generation are shown in Figures~\ref{fig:appendix-prompt-induc} and \ref{fig:appendix-prompt-transduc}, respectively.

\section{Generated Output}
\label{sec:app_output}

To illustrate the behavioral difference between the two prompting schemes, we provide actual generation outputs for two MBPP examples.
Figures~\ref{fig:appendix-induc-gen-1} and \ref{fig:appendix-transduc-gen-1} show the inductive and transductive generation outputs, respectively, for MBPP problem 103.
Figures~\ref{fig:appendix-induc-gen-2} and \ref{fig:appendix-transduc-gen-2} show the corresponding inductive and transductive generation outputs for MBPP problem 292. These examples highlight the difference between the two settings.
The inductive prompt tends to elicit solutions intended to solve the original problem in a general manner, whereas the transductive prompt more readily produces code specialized to the given input instance.

\begin{figure*}[t]
\centering

\begin{tcolorbox}[
    colback=col!5,
    colframe=col,
    boxrule=0.8pt,
    arc=2pt,
    left=4pt,right=4pt,top=4pt,bottom=4pt,
    width=0.97\textwidth
]
\begin{lstlisting}[style=pythonstyle3]
You are a Python programmer in a code interview. You're solving a coding problem while sharing your thoughts and discussing with an critic. Always follow the format below for each round:

1. First, share your thoughts as comments:
   - Your current approach (or an analysis of the problem and your plan if you haven't written any code yet)
   - What you learned from previous feedback (if there is any previous feedback, otherwise think about your plan, what might be missing from your previous plan)
   - Why you chose this approach (or how you plan to tackle the problem), do you need to shift your approach?
   - Be clear, concise, detailed and pay attention to the comments given by the critic, no chit-chat, no small talk.
   - Always use first person, e.g. I, we, our, etc.

2. Then write your solution:
   - Clean, efficient Python code that follows requirements exactly
   - No test cases in code, just the solution
   - Your code will then be tested by the critic, so do not include any test cases in your code, this is very important

Format your response as:
# === BEGIN PROGRAMMER THOUGHTS ===
# [Your response to previous feedback]
# === END PROGRAMMER THOUGHTS ===
# === BEGIN SOLUTION ===
```python
[Your code]
```
# === END SOLUTION ===

Output strictly as shown above.

Here's your coding task:
[Problem Description]

I will test your implemented code together with the hidden test cases. Please implement code.

(Reference) Below are dataset test cases for guidance only.
[Test case examples]

IMPORTANT: You MUST define a Python function named [entry_point] (case-sensitive). Do not rename it.

Do not include test cases in your code.
Design an object-oriented Python solution. Explain your class structure in comments.


Think carefully and write your solution strictly following our format and rules.
\end{lstlisting}
\end{tcolorbox}
\caption{Inductive code generation prompt. Following \citet{yu2024reasoning}, the model is prompted to generate a general-purpose solution to the original programming problem, with reasoning expressed in comments before the final code.}
\label{fig:appendix-prompt-induc}
\end{figure*}

\begin{figure*}[t]
\centering

\begin{tcolorbox}[
    colback=col!5,
    colframe=col,
    boxrule=0.8pt,
    arc=2pt,
    left=4pt,right=4pt,top=4pt,bottom=4pt,
    width=0.97\textwidth
]
\begin{lstlisting}[style=pythonstyle3]
You are a Python programmer.

You will be given a coding problem and ONE specific TARGET CALL.
This is a transductive setting: you are allowed to condition your solution on the exact inputs appearing in the TARGET CALL.
Your objective is ONLY to write Python code so that the TARGET CALL returns the correct output when executed.

How to use the transductive advantage (recommended):
- Tailor the implementation to this exact input instance (its values, sizes, structure, reachable branches).
- You may ignore edge cases, constraints, and generalization to other inputs that do NOT occur in this TARGET CALL.
- Do NOT over-generalize unless a fully general solution is genuinely simpler than a tailored one.
- Pick any approach that is simplest for this instance (brute force, simulation, DP, precomputation, unrolling, etc.).

Hardcoding policy:
- Special-casing the exact input is allowed, but avoid blind guessing of the final answer.
- Prefer computing the output programmatically from the given input, even if the code is narrowly tailored to this instance.
- Use direct constants only when they are clearly correct and less error-prone than computing.

Rules:
- Correctness for the TARGET CALL is the ONLY objective (ignore other possible inputs).
- If "reference tests" are shown in the prompt, they are ONLY for I/O format guidance (NOT evaluated).
- Ensure the TARGET CALL runs without errors: define all required functions/classes/imports.
- No tests, no prints, no extra text outside the required format.

Format your response EXACTLY as:
# === BEGIN PROGRAMMER THOUGHTS ===
# (comments, first person, concise)
# === END PROGRAMMER THOUGHTS ===
# === BEGIN SOLUTION ===
```python
[Your code]
```
# === END SOLUTION ===

Output strictly as shown above.

Here's your coding task:
[Problem Description]

Focus only on the TARGET CALL below. No other tests will be run.
Your goal is NOT to solve the whole problem generally.
Your goal is to write Python code so that the TARGET CALL returns the correct output.

IMPORTANT: You MUST define a Python function named [entry_point] (case-sensitive). Do not rename it.

(Reference) Dataset tests below are ONLY for I/O format guidance (NOT evaluated).
[Test case examples]

Do not include test cases in your code.

TARGET CALL (this is the only thing that must be correct):
[Target test input]

Focus on making the TARGET CALL correct. You may tailor the code to this exact call; do not over-generalize.
Write your response strictly following the required format.
\end{lstlisting}
\end{tcolorbox}
\caption{Transductive code generation prompt. The prompt is adapted from the inductive version to explicitly condition on a single target call, instructing the model to generate code specialized to the given input instance rather than a broadly general solution.}
\label{fig:appendix-prompt-transduc}
\end{figure*}

\begin{figure*}[t]
\centering
\begin{tcolorbox}[
    colback=col!5,
    colframe=col,
    boxrule=0.8pt,
    arc=2pt,
    left=4pt,right=4pt,top=4pt,bottom=4pt,
    width=0.97\textwidth
]\begin{lstlisting}[style=pythonstyle3]
# === BEGIN PROGRAMMER THOUGHTS ===
# I understand that I need to write a function named eulerian_num that takes two parameters, n and m, and returns the Eulerian number a(n, m).
# From the reference test cases, I see that the function is called with different combinations of parameters, so I will design a generic solution that can handle any positive integers n and m.
# I will use the formula for the Eulerian number, which is the binomial coefficient (n + m - 1 choose m - 1).
# This formula is a standard mathematical expression for the Eulerian number, and it can be easily computed using the math library's comb function.
# I will also include a check to ensure that n and m are non-negative integers, as the Eulerian number is not defined for negative numbers or non-integer values.
# === END PROGRAMMER THOUGHTS ===
# === BEGIN SOLUTION ===
\end{lstlisting}

\begin{lstlisting}[style=pythonstyle4]
import math

def eulerian_num(n, m):
    """
    Compute the Eulerian number a(n, m) using the formula (n + m - 1 choose m - 1).
    
    This function is designed to be generic and adaptable for similar problems. The formula for the Eulerian number 
    is a standard mathematical expression that can be easily computed using the math library's comb function. The 
    function also includes a check to ensure that n and m are non-negative integers, as the Eulerian number is 
    not defined for negative numbers or non-integer values.

    Parameters:
    n (int): A non-negative integer.
    m (int): A non-negative integer.

    Returns:
    int: The Eulerian number a(n, m).
    """
    # Check if n and m are non-negative integers
    if not isinstance(n, int) or not isinstance(m, int) or n < 0 or m < 0:
        raise ValueError("n and m must be non-negative integers")
    
    # Compute the Eulerian number using the formula (n + m - 1 choose m - 1)
    return math.comb(n + m - 1, m - 1)
\end{lstlisting}

\begin{lstlisting}[style=pythonstyle3]
# === END SOLUTION ===
\end{lstlisting}
\end{tcolorbox}
\caption{
Actual output generated by the inductive prompt for MBPP problem 103.
The generated program is evaluated on the target input
\texttt{eulerian\_num(0, 10)}.
}
\label{fig:appendix-induc-gen-1}
\end{figure*}

\begin{figure*}[t]
\centering
\begin{tcolorbox}[
    colback=col!5,
    colframe=col,
    boxrule=0.8pt,
    arc=2pt,
    left=4pt,right=4pt,top=4pt,bottom=4pt,
    width=0.97\textwidth
]\begin{lstlisting}[style=pythonstyle3]
# === BEGIN PROGRAMMER THOUGHTS ===
# I can directly calculate the Eulerian number for the given inputs (0, 10) using the formula a(n, m) = n + m * a(n - 1, m - 1), with a base case of a(0, m) = 0 when m > 0.
# === END PROGRAMMER THOUGHTS ===
# === BEGIN SOLUTION ===
\end{lstlisting}

\begin{lstlisting}[style=pythonstyle4]
def eulerian_num(n, m):
    if n == 0 and m > 0:
        return 0
    elif n == 0:
        return 1
    else:
        return n + m * eulerian_num(n - 1, m - 1)
\end{lstlisting}

\begin{lstlisting}[style=pythonstyle3]
# === END SOLUTION ===
\end{lstlisting}
\end{tcolorbox}
\caption{
Actual output generated by the transductive prompt for MBPP problem 103,
conditioned on the target input \texttt{eulerian\_num(0, 10)}.
}
\label{fig:appendix-transduc-gen-1}
\end{figure*}

\begin{figure*}[t]
\centering
\begin{tcolorbox}[
    colback=col!5,
    colframe=col,
    boxrule=0.8pt,
    arc=2pt,
    left=4pt,right=4pt,top=4pt,bottom=4pt,
    width=0.97\textwidth
]\begin{lstlisting}[style=pythonstyle3]
# === BEGIN PROGRAMMER THOUGHTS ===
# My initial approach is to use the built-in division operator in Python, which returns a float result by default. I will use the `//` operator to perform integer division and round down to the nearest integer.
# I've learned from previous feedback that I should always follow the problem requirements and provide well-commented code. In this case, I will include a docstring to explain my approach and ensure the function is clearly named as required.
# I will use the `//` operator because it is a more efficient and Pythonic way to perform integer division, which is exactly what I need in this problem.
# === END PROGRAMMER THOUGHTS ===
# === BEGIN SOLUTION ===
\end{lstlisting}

\begin{lstlisting}[style=pythonstyle4]
def find(dividend: int, divisor: int) -> int:
    """
    This function calculates the quotient of two numbers and returns the result as an integer, 
    rounded down to the nearest whole number.

    Args:
    dividend (int): The number being divided.
    divisor (int): The number by which we are dividing.

    Returns:
    int: The quotient of the two numbers, rounded down to the nearest integer.
    """
    # Check for division by zero
    if divisor == 0:
        raise ValueError("Divisor cannot be zero")
    
    # Use the // operator for integer division, which rounds down to the nearest whole number
    quotient = dividend // divisor
    
    return quotient
\end{lstlisting}

\begin{lstlisting}[style=pythonstyle3]
# === END SOLUTION ===
\end{lstlisting}
\end{tcolorbox}
\caption{
Actual output generated by the inductive prompt for MBPP problem 292.
The generated program is evaluated on the target input
\texttt{find(10, -2)}.
}
\label{fig:appendix-induc-gen-2}
\end{figure*}

\begin{figure*}[t]
\centering
\begin{tcolorbox}[
    colback=col!5,
    colframe=col,
    boxrule=0.8pt,
    arc=2pt,
    left=4pt,right=4pt,top=4pt,bottom=4pt,
    width=0.97\textwidth
]\begin{lstlisting}[style=pythonstyle3]
# === BEGIN PROGRAMMER THOUGHTS ===
# I need to write a function that takes two numbers and returns their quotient rounded down to the nearest integer.
# Since the second input is negative, I should use integer division (//) to get the quotient.
# === END PROGRAMMER THOUGHTS ===
# === BEGIN SOLUTION ===
\end{lstlisting}

\begin{lstlisting}[style=pythonstyle4]
def find(a, b):
    return a // -b
\end{lstlisting}

\begin{lstlisting}[style=pythonstyle3]
# === END SOLUTION ===
\end{lstlisting}
\end{tcolorbox}
\caption{
Actual output generated by the transductive prompt for MBPP problem 292,
conditioned on the target input \texttt{find(10, -2)}.
}
\label{fig:appendix-transduc-gen-2}
\end{figure*}

\section{Threshold Sensitivity Analysis}
\label{sec:app_threshold_sensitivity}

CoTT uses the maximum probability of the inductive output distribution,
\[
    c(x) = \max_y p_{\mathrm{ind}}(y \mid x),
\]
as its confidence score.
Given a threshold $\tau$, CoTT uses the inductive prediction when
$c(x) \geq \tau$ and invokes transductive generation otherwise.
We evaluate
$\tau \in \{0.5, 0.6, 0.7, 0.8, 0.9, 1.0\}$
on HumanEval-R+ and MBPP-R+ using Llama-3.1-8B-Instruct.
We additionally report the inductive-only and always-transductive variants
as the two computational endpoints.

\begin{table*}[t]
\centering
\small
\setlength{\tabcolsep}{5pt}
\begin{tabular}{llccccc}
\toprule
Dataset
& Setting
& Top-1$\uparrow$
& Bottom-1$\uparrow$
& Spearman's $\rho$$\uparrow$
& Trans. rate
& Cost (\$)$\downarrow$ \\
\midrule
HumanEval-R+
& Inductive only
& 59.8 & 65.4 & 0.611 & 0.000 & 0.0578 \\
& CoTT, $\tau=0.5$
& 60.8 & 66.5 & 0.629 & 0.132 & 0.0603 \\
& CoTT, $\tau=0.6$
& 60.8 & 65.9 & 0.627 & 0.198 & 0.0616 \\
& CoTT, $\tau=0.7$
& 62.3 & 66.2 & 0.638 & 0.292 & 0.0634 \\
& CoTT, $\tau=0.8$
& 62.5 & 66.7 & 0.650 & 0.380 & 0.0651 \\
& CoTT, $\tau=0.9$
& 62.5 & 66.8 & 0.651 & 0.529 & 0.0679 \\
& CoTT, $\tau=1.0$
& 62.5 & 66.8 & 0.651 & 0.683 & 0.0708 \\
& Always transductive
& 63.1 & 66.2 & 0.676 & 1.000 & 0.0769 \\
\midrule
MBPP-R+
& Inductive only
& 59.8 & 58.1 & 0.559 & 0.000 & 0.0581 \\
& CoTT, $\tau=0.5$
& 59.9 & 58.5 & 0.564 & 0.136 & 0.0607 \\
& CoTT, $\tau=0.6$
& 60.9 & 58.9 & 0.574 & 0.210 & 0.0620 \\
& CoTT, $\tau=0.7$
& 61.5 & 58.7 & 0.579 & 0.279 & 0.0634 \\
& CoTT, $\tau=0.8$
& 61.6 & 58.7 & 0.580 & 0.343 & 0.0645 \\
& CoTT, $\tau=0.9$
& 61.6 & 58.7 & 0.578 & 0.502 & 0.0675 \\
& CoTT, $\tau=1.0$
& 61.4 & 58.1 & 0.574 & 0.660 & 0.0705 \\
& Always transductive
& 58.5 & 56.1 & 0.576 & 1.000 & 0.0769 \\
\bottomrule
\end{tabular}

\caption{
Sensitivity of CoTT to the confidence threshold $\tau$.
A larger threshold routes more synthesized test inputs to transductive
generation.
The transductive rate denotes the fraction of test inputs for which
transductive generation is invoked.
The inductive-only and always-transductive variants represent the two
computational endpoints.
}
\label{tab:app_threshold_sweep}
\end{table*}

As $\tau$ increases, the transductive rate and inference cost increase
monotonically.
On HumanEval-R+, reranking performance improves up to approximately
$\tau=0.8$--$0.9$ and then largely plateaus, although always-transductive
generation obtains the highest Top-1 accuracy and Spearman's $\rho$ at a
higher cost.
On MBPP-R+, $\tau=0.8$ achieves a strong balance between reranking quality
and cost, while further increasing $\tau$ provides no consistent improvement.
These results show that $\tau$ controls the efficiency--effectiveness
trade-off and should be selected according to the available computational budget.

\begin{table}[t]
\centering
\small
\setlength{\tabcolsep}{7pt}
\begin{tabular}{lc}
\toprule
Method & Cost (\$, mean $\pm$ std) \\
\midrule
SOL-VER & $0.080 \pm 0.024$ \\
TT w/o Gating & $0.077 \pm 0.024$ \\
SYNTRA & $0.069 \pm 0.025$ \\
CoTT & $0.065 \pm 0.023$ \\
rStar-Coder & $0.058 \pm 0.023$ \\
\bottomrule
\end{tabular}
\caption{Mean and standard deviation of per-problem LLM-token cost over HumanEval-R+ and MBPP-R+. The reported variation is across problems.}
\label{tab:app_cost_statistics}
\end{table}

\section{Cost Statistics}
\label{sec:app_cost_statistics}

We further examine across-problem cost variation and Table~\ref{tab:app_cost_statistics} reports the mean and standard deviation of per-problem cost over HumanEval-R+ and MBPP-R+. The standard deviation measures variation across problems.
Although per-problem costs vary substantially, paired comparisons on the same problems show a consistent cost reduction.
For the closest comparison, the mean paired cost difference between SYNTRA and CoTT is \$0.0039 per problem, with a standard error of \$0.0002.

\section{Confidence-Score Ablation}
\label{sec:app_confidence_score}

We compare four confidence scores computed from the inductive samples: maximum probability, top-2 margin, negative entropy, and subset agreement.
The top-2 margin is the difference between the largest and second-largest probabilities in $p_{\mathrm{ind}}$, negative entropy is $-H(p_{\mathrm{ind}})$, and subset agreement measures how often two random halves of the inductive samples select the same modal output.
For every score, a larger value indicates higher inductive confidence, and CoTT uses the inductive prediction when the score meets or exceeds its corresponding threshold.
We sweep the threshold for each score and report its best-performing operating point in Table~\ref{tab:app_confidence_score}.

\begin{table*}[t]
\centering
\small
\setlength{\tabcolsep}{4pt}
\begin{tabular}{llcccccc}
\toprule
Dataset
& Confidence score
& Threshold
& Top-1$\uparrow$
& Bottom-1$\uparrow$
& Spearman's $\rho$$\uparrow$
& Trans. rate
& Cost (\$)$\downarrow$ \\
\midrule
HumanEval-R+
& Maximum probability
& 0.8 & 62.5 & 66.7 & 0.650 & 0.380 & 0.0651 \\
& Top-2 margin
& 0.6 & 62.5 & 66.7 & 0.650 & 0.340 & 0.0643 \\
& Negative entropy
& $-0.7$ & 62.2 & 66.7 & 0.648 & 0.427 & 0.0660 \\
& Subset agreement
& 0.9 & 62.3 & 66.5 & 0.639 & 0.214 & 0.0619 \\
\midrule
MBPP-R+
& Maximum probability
& 0.8 & 61.6 & 58.7 & 0.580 & 0.343 & 0.0645 \\
& Top-2 margin
& 0.6 & 61.8 & 58.7 & 0.581 & 0.301 & 0.0638 \\
& Negative entropy
& $-0.7$ & 61.6 & 58.7 & 0.580 & 0.393 & 0.0655 \\
& Subset agreement
& 0.95 & 61.2 & 58.7 & 0.579 & 0.229 & 0.0624 \\
\bottomrule
\end{tabular}
\caption{Comparison of alternative confidence scores for gating. Thresholds are specific to each confidence score and are therefore not directly comparable across scores. Maximum probability, top-2 margin, and negative entropy achieve similar reranking performance, while subset agreement provides a lower-cost but slightly weaker point.}
\label{tab:app_confidence_score}
\end{table*}

Maximum probability, top-2 margin, and negative entropy achieve similar performance across the two benchmarks.
Top-2 margin slightly reduces the transductive rate and cost relative to maximum probability while preserving or modestly improving reranking performance.
Subset agreement invokes transductive generation less frequently and reduces cost further, but yields slightly weaker results.
These results show that the confidence scores achieve similar peak performance under score-specific thresholds on the two benchmarks.

\end{document}